\documentclass[runningheads]{llncs}
\usepackage[T1]{fontenc}
\usepackage{graphicx}
\usepackage{subcaption}
\usepackage{multirow}
\usepackage{mdframed}
\usepackage{listings}
\usepackage{dirtree}
\usepackage{booktabs}
\usepackage{comment}
\usepackage{enumitem}
\usepackage{xcolor}
\usepackage{booktabs}
\usepackage{hyperref}
\usepackage{wrapfig}
\renewcommand{\paragraph}[1]{\ \\[-3mm]\noindent\textbf{#1}}

\begin{document}
\title{Refining Wikidata Taxonomy using Large Language Models}
%
%
\author{Yiwen Peng\inst{1}\orcidID{0009-0007-7902-4097} \and
Thomas Bonald\inst{1}\orcidID{0000-0003-0468-0384} \and
Mehwish Alam\inst{1}\orcidID{0000-0002-7867-6612}}
\authorrunning{Y. Peng et al.}
%
\institute{Télécom Paris, Institut Polytechnique de Paris, France\\
\email{\{first.last\}@telecom-paris.fr}}
%
\maketitle              
\begin{abstract}
  Due to its collaborative nature, Wikidata is known to have a complex taxonomy, with recurrent issues like the ambiguity between instances and classes, the inaccuracy of some taxonomic paths, the presence of cycles, and the high level of redundancy across classes. Manual efforts to clean up this taxonomy are time-consuming and prone to errors or subjective decisions. We present WiKC, a new version of Wikidata taxonomy cleaned automatically using a combination of Large Language Models (LLMs) and graph mining techniques. Operations on the taxonomy, such as cutting links or merging classes,  are performed with the help of zero-shot prompting on an open-source LLM. The quality of the refined taxonomy is evaluated from both intrinsic and extrinsic perspectives, on a task of entity typing for the latter, showing the practical interest of WiKC.
  
\keywords{Knowledge Graphs \and Large Language Models \and Graph Mining \and Taxonomy Refinement.}
\end{abstract}
\section{Introduction}

Wikidata is a general-purpose Knowledge Base (KB) maintained by a large community of contributors. 
As a collaborative project, Wikidata faces several challenges, including the ambiguity, inconsistency, redundancy, and complexity of its taxonomy.
Ambiguity arises from the confusion between instances and classes. For example, {\it scientist} (Q901) is both an instance of {\it profession} (Q28640) and a subclass of {\it person} (Q215627). 
Inconsistency here refers to the inaccuracy of some taxonomic paths. For instance,  {\it city} (Q515) is inaccurately classified as a subclass of {\it mathematical object} (Q246672) through the following taxonomic path: {\it city} (Q515) → {\it spatial entity} (Q58416391) → {\it geometric object} (Q123410745) → {\it mathematical object} (Q246672). 
Redundancy is also prevalent with classes like {\it human} (Q5) and {\it person} (Q215627) coexisting, where one would suffice. 
The complexity of the taxonomy is another major issue. The taxonomy of Wikidata has a depth of 20, contains many cycles, like {\it axiom} (Q17736) $\to$ {\it first principle} (Q536351) $\to$ {\it principle} (Q211364) $\to$ {\it axiom} (Q17736), and transitive links, like {\it airport} (Q1248784) → {\it aerodrome} (Q62447) → {\it station} (Q12819564) and {\it airport} (Q1248784) → {\it station} (Q12819564). Additionally, only 4\% of the 4 million classes are instantiated, with many lacking labels and descriptions.

In this paper, we propose an approach for refining Wikidata taxonomy and thus addressing some of the above issues. Unlike YAGO 4.5 \cite{suchanek2024yago}, where the upper taxonomy of Wikidata is manually mapped to Schema.org\footnote{\url{https://schema.org}} while the lower taxonomy remains unchanged, we adopt an automated approach for refining the whole taxonomy, based on a combination of Large Language Models (LLMs) and graph mining techniques. 
Specifically, we use a zero-shot prompt on each link of the taxonomy, asking the LLM to predict one of the following relations: {\it subclassOf}, {\it superclassOf}, {\it equivalent}, {\it irrelevant}, or {\it none}. 
Given different predictions, we decide whether to cut the link, merge the classes, or keep the classes and their link unchanged. This yields  WiKC, a cleaned version of  Wikidata taxonomy, which we make publicly available\footnote{\url{https://github.com/peng-yiwen/WiKC}}. 
The quality of WiKC is then evaluated from both intrinsic and extrinsic perspectives. For extrinsic evaluation, we design a judge LLM on the task of entity typing to compare WiKC with the original taxonomy of Wikidata. For the sake of reproducibility, both steps (refinement and evaluation) are based on an open-source LLM.

The paper is structured as follows. Section \ref{sec:related_work} covers the related work, Section \ref{sec:approach} describes our approach for taxonomy refinement, Section \ref{sec:evaluation} presents the evaluation 
and Section \ref{sec:conclude} concludes the paper.

\section{Related Work \label{sec:related_work}}

\paragraph{General-purpose knowledge bases.} 
Wikidata \cite{vrandevcic2014wikidata} is the largest open general-purpose KB, maintained by a large community of contributors. Due to its collaborative nature, Wikidata is known to have a complex taxonomy, including errors, redundancies and inconsistencies
\cite{shenoy2022study,brasileiro2016applying}.
Cleaning this taxonomy is the main objective of our work.
Other general-purpose KBs include DBpedia \cite{auer2007dbpedia} and YAGO 4.5 \cite{suchanek2024yago}.
DBpedia is a multilingual KB automatically extracted from Wikipedia and more recently,  Wikidata. Its ontology covers a wide range of concepts 
but suffers from inconsistencies \cite{dbpediaInconsist2012,abian2018inconsist} due to its reliance on Wikipedia and the prioritization of coverage over precision. 
YAGO 4.5  is based on Wikidata, with a manual mapping of 
 the upper taxonomy to Schema.org, providing a clean upper-level taxonomy designed by human experts. 
In this paper, we propose an automatic approach for refining the Wikidata taxonomy, without requiring any human expertise or subjective decision. 

\paragraph{Taxonomy refinement, taxonomy induction.} 
Taxonomy refinement is the task of updating an existing taxonomy while maintaining its structure. Previous methods are either domain-specific \cite{o2017restructuring} or depend on lexical structures of existing hierarchies \cite{ponzetto2009large}. 
Recently, more advanced approaches have incorporated word embeddings into taxonomy refinement. For instance,   hyperbolic embeddings are used in \cite{aly2019every} to detect outliers in a domain-specific taxonomy. In \cite{malandri2021taxoref}, a hierarchical semantic similarity metric is used to select better embeddings and then refine a taxonomy. 

Few studies have explored the use of LLMs in taxonomy refinement. Instead, LLMs have been applied to the closely related task of taxonomy induction \cite{babaei2023llms4ol,chen2023prompting,funk2023towards,zeng2024chain}, which derives a taxonomy from scratch given entities extracted from text. For example, an approach called Chain-of-Layer is used in \cite{zeng2024chain} to select relevant entities. A zero-shot knowledge-agnostic strategy is used in \cite{carta2023iterative}  for constructing the upper levels of a taxonomy. In \cite{funk2023towards},   a concept hierarchy is generated for a given domain starting from a seed concept. 
  The reliability of LLMs in hierarchical structure discovery is demonstrated in \cite{sun2024large} for common knowledge graphs, including Schema.org.
While these studies show the potential of LLMs in taxonomy induction, the automatic refinement of a large common taxonomy like that of Wikidata remains an open challenge.

\section{Approach \label{sec:approach}}

In our work, we use the {\it truthy} version of Wikidata\footnote{Data dump dated March 22, 2024, and 949GB after unpacking.}, which contains the best non-deprecated rank for each property.

\subsection{Taxonomy Extraction \label{sec:taxon_extract}}

In principle, the taxonomy of Wikidata is defined by the  {\it subclassOf} (P279) property. In practice, this property is often confused with  the 
 {\it instanceOf} (P31) property by contributors to Wikidata, requiring some work to extract the actual taxonomy. 

\textbf{Instance or class?} We extract instances and classes using the {\it instanceOf} and  {\it subclassOf} properties, respectively, giving priority to the {\it instanceOf} property if both appear. For example, the entity {\it hydrogen} (Q556), which is both an instance of {\it chemical element} (Q11344) and a subclass of  {\it energetic material} (Q5376832), is considered as an instance, not a class. However, some exceptions must be taken into account. For example, the entity {\it company} (Q783794), which is an instance of  {\it type of organisation} (Q17197366) and a subclass of {\it organization} (Q43229), should be considered as a class. The difference with the previous example is that the entity {\it type of organisation} is, in fact, a {\it metaclass} (Q19478619), i.e., {\it a class which has instances that are all themselves classes}. Given that,  the entity {\it company}  should indeed be considered as a class, not an instance. 

In our case, we consider that an entity is a meta-class if it is an instance of either {\it metaclass} (Q19478619) or {\it second-order class} (Q24017414) that meets the following criteria: (1) Its label contains a keyword like {\it type}, {\it class}, {\it style}, {\it genre}, {\it form}, {\it occupation}, {\it profession}, {\it category}, {\it classification}, (2) Its label does not contain a preposition, which corresponds to very specific classes, 
nor the keyword {\it property}, which refers to classes of properties. 

We also exclude {\it BFO class} (Q124711104) from the meta-classes to avoid external ontologies. We finally obtain 434 meta-classes and approximately 1.7M classes (either an instance of a meta-class or an entity that has the {\it subclassOf} property and not the {\it instanceOf}  property; meta-classes are excluded). 
The class {\it product} (Q2424752) was manually added because it is an important class, e.g., one of the 11 top-level classes of Schema.org, filtered out by the previous process as an instance of  {\it economic concept} (Q29028649).

\textbf{Graph construction.} Among all entities declared as classes by the previous process, we only keep those with a label. We build a directed graph between these 1.6M classes using the  {\it subclassOf} property.  
We then explore the taxonomy from the root class {\it entity} (Q35120)  by depth-first-search (DFS) and break any link that would create a cycle. During traversal, we also bypass any class without a description, as most of these classes are either redundant or overly specific. For example, the class {\it award for best screenplay} (Q96474700) is a subclass of {\it award for best book} (Q105810971), which is itself a subclass of {\it literary award} (Q378427); here we bypass {\it award for best book} due to the absence of description in Wikidata.

\textbf{Filtering.} We exclude the class {\it scholarly article} (Q13442814) and all its successors in the graph, as the addition of scholarly articles to Wikidata has been controversial\footnote{\url{https://www.mail-archive.com/wikidata@lists.wikimedia.org/msg06716.html}} and would contribute more than 40M entities. We also remove classes that do not have instances (there are 252k of them). In addition, we eliminate top-level classes (subclasses of root class {\it entity}) that do not have subclasses themselves, such as {\it unidentified entity} (Q120725535) and {\it named entity} (Q25047676). These classes are either derived from external ontologies or are too general to be useful in our taxonomy.

After completing the aforementioned steps, we obtain an acyclic taxonomy with approximately 40k classes and 53k links, where each class has a label and a description.

\subsection{Taxonomy Refinement \label{sec:taxon_refine}}

\begin{figure*}[htbp]
  \centering
  \begin{subfigure}[b]{0.32\linewidth}
    \includegraphics[width=\linewidth]{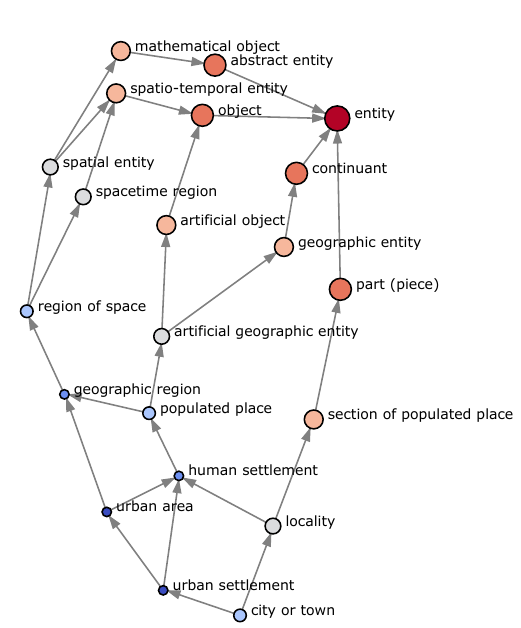}
    \caption{\small Original}
    \label{figure:original}
  \end{subfigure}
  \begin{subfigure}[b]{0.29\linewidth}
    \includegraphics[width=\linewidth]{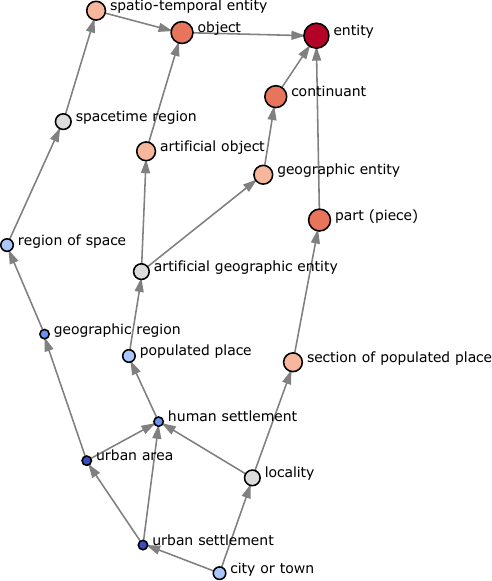}
    \caption{\small Cut}
    \label{figure:cut}
  \end{subfigure} \\
  \vspace{0.5cm}
  \begin{subfigure}[b]{0.25\linewidth}
    \includegraphics[width=\linewidth]{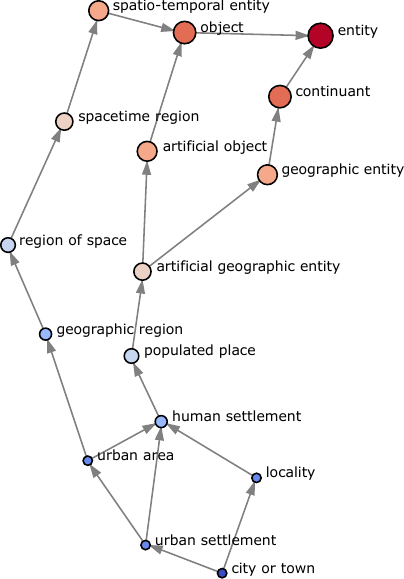}
    \caption{Resolve}
    \label{figure:resolve}
  \end{subfigure}
  \hspace{0.5cm}
  \begin{subfigure}[b]{0.25\linewidth}
    \includegraphics[width=\linewidth]{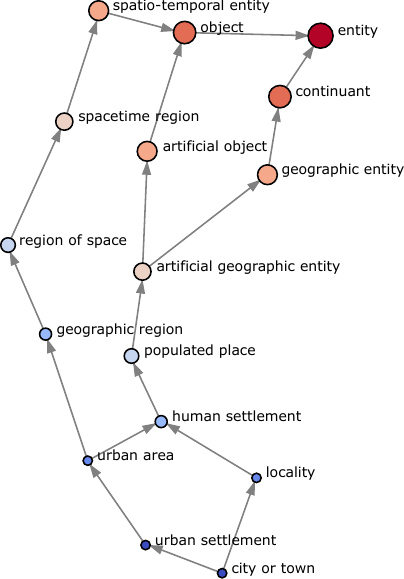}
    \caption{Reduce}
    \label{figure:reduce}
  \end{subfigure} \\ 
  \vspace{0.5cm}
  \begin{subfigure}[b]{0.25\linewidth}
    \includegraphics[width=\linewidth]{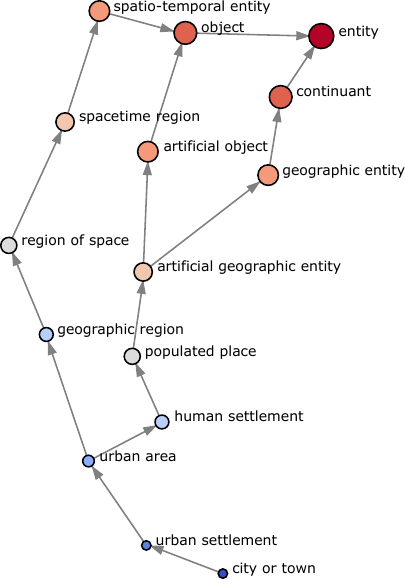}
    \caption{Merge \& \small rewire}
    \label{figure:merge}
  \end{subfigure}
  \hspace{0.5cm}
  \begin{subfigure}[b]{0.25\linewidth}
    \includegraphics[width=\linewidth]{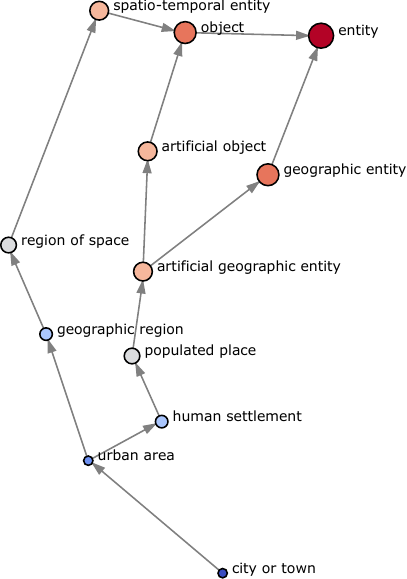}
    \caption{Filter}
    \label{figure:final}
  \end{subfigure}
  \caption{ \label{figure:city} Taxonomy from {\it city or town} {\rm (Q7930989)} to  {\it entity} {\rm (Q35120)} after each step of the refinement. }
\end{figure*}

We refine the taxonomy to address issues like redundancy and inconsistency. For instance,  the class {\it city or town} (Q27676416) is a subclass of  {\it city or town} (Q7930989) (redundancy) and a transitive subclass of 
{\it mathematical object} (Q246672) (inconsistency). 
For this, we prompt an LLM (see details in \S\ref{llm}) to analyze each link of the graph and predict the correct semantic relation from the following ones: {\it subclassOf}, {\it superclassOf}, {\it equivalent}, {\it irrelevant}, and {\it none}. The {\it superclassOf} prediction is used to potentially reverse the link direction.  Given these results, obtained for each link of the graph, we apply the following steps sequentially: (1) \textbf{Cut} irrelevant links; (2) \textbf{Resolve} reversed links; (3) \textbf{Reduce} transitive links; (4) \textbf{Merge}  equivalent classes; (5) \textbf{Rewire} links upon confirmation by LLM and (6) \textbf{Filter} out non-informative and rare classes.
The evolution of the subgraph corresponding to the paths from {\it city or town} (Q7930989) to {\it entity} (Q35120) is shown in Figure \ref{figure:city},  after each refining step. The details for each step are described below. 

\textbf{Cut.} 
Any link that is predicted as {\it irrelevant} or {\it none} is cut if the corresponding classes remain connected to the root class {\it entity} after the cut, or if the disconnected subgraph after the cut has at most 3 nodes (in which case the corresponding classes are also removed from the taxonomy). Links are considered in order of distance from the root class {\it entity}.
In Figure \ref{figure:cut} for instance, {\it city or town} is no longer a transitive subclass of {\it mathematical object}, because the link from {\it spatial entity} to {\it mathematical object} is cut.

\textbf{Resolve.} 
When the link prediction is reversed, i.e., class $A$ is predicted as a {\it superclassOf}  class $B$ instead of a {\it subclassOf} class $B$, we merge these classes or cut their link depending on the connectivity of class $A$ in the graph: if class $A$ is a subclass of other classes than $B$, the link between classes $A$ and $B$ is cut; otherwise, classes $A$ and $B$ are merged.
Merging is reasonable in this resolving step as reversed links are usually caused by similar classes, making it hard to decide which one is a subclass of the other. 
In Figure \ref{figure:resolve}, the link from {\it locality}  to {\it section of populated place}  is cut as {\it locality} has another superclass, {\it human settlement}. 

\textbf{Merge.} We merge classes that have exactly the same label or that are predicted as equivalent by the LLM. 
Specifically, if class $A$ is predicted as  {\it equivalent} to class $B$ instead of a {\it subclassOf}  class $B$, class $A$ is merged into class $B$.  The subclasses of class $A$ are then relinked to class $B$ as its subclasses.  A transitive reduction is performed after each merging operation to prevent the introduction of transitive links and cycles in the taxonomy. 
In Figure \ref{figure:merge} for instance, the class {\it locality}, defined as a {\it place of human settlement}, is merged into class {\it human settlement} as these two classes are predicted as equivalent.

\textbf{Rewire.} After the previous merge step, we inspect the potential subclass links between class $B$ (the former superclass of class $A$), and the other superclasses of class $A$, if any.
Here we use the same LLM and prompt to check these links and only accept those correctly predicted as {\it subclassOf}.  Our experiments show that, out of such 280 potential subclass links, 118 are correctly predicted by the LLM. For instance, after the merge of {\it tweezers} (Q192504)  into {\it forceps} (Q1378235),  {\it forceps} becomes a subclass of  {\it hand tool} (Q2578402),  another superclass of {\it tweezers}, after confirmation of the LLM.

\textbf{Filter.} 
In this final step, to further overcome redundancy issues, we remove recursively {\it non-informative} classes (classes with only one superclass, one subclass, and without direct instances), {\it rare} classes (classes with at most one instance including both direct and transitive instances, or without a Wikipedia page\footnote{For classes without any Wikipedia page, we consider only classes with a depth higher than 3, as the upper-level classes might be too abstract to be described in Wikipedia  (e.g., {\it artificial object} (Q16686448)).}), 
and {\it specific} top-level classes (top-level classes whose subclasses are all linked to other deeper level classes)\footnote{For instance, 
{\it testbed} (Q1318674) is a \textit{specific} top-level class. The subclass of {\it testbed}, {\it elevator test tower} (Q1689156), is also a subclass of a deeper level class {\it tower} (Q12518), and {\it testbed} is too specific to be directly linked to the root class {\it entity} (Q35120).}.
The remaining classes are reconnected respecting the previous taxonomy structure if no transitive links are created. 
In Figure \ref{figure:final} for instance, the non-informative class {\it spacetime region}  is removed and the class {\it region of space} becomes a subclass of {\it spatio-temporal entity}; the rare class {\it urban settlement}  is also removed (due to the absence of a Wikipedia page) and the class {\it city or town} becomes a subclass of {\it urban area}.

After refining the taxonomy of Wikidata, we obtain a new acyclic taxonomy called WiKC, with approximately 17k classes and 20k links, without any cycles and transitive links.

\subsection{Large Language Models \label{llm}}

For the sake of reproducibility, we use the open-source LLM \textit{Mixtral-8x7B-Instruct-v0.1}\footnote{\url{https://huggingface.co/mistralai/Mixtral-8x7B-Instruct-v0.1}}, with a temperature set to zero to get deterministic results.
We formulate a prompt to enable the LLM to generate answers from the context, including class labels and descriptions. Inspired by the chain-of-thought \cite{wei2022chain}, which bridges the reasoning gap between input and answer, we add an explanation part before the answer to ensure careful analysis before predicting the semantic relation. The corresponding prompt is shown below. 

\noindent\makebox[0pt][l]{
\colorbox{blue!6}{
  \parbox{11.5cm}{
    \scriptsize
    \texttt{
    \textbf{\%Instructions:}\\
      You are a linguistic expert who understands the semantic relationships between concepts. 
      Your task is to determine the most appropriate semantic relation between two provided concepts based on the available labels and descriptions. 
      The potential relations are: "subclass of", "superclass of", "equivalent to", "irrelevant to", or "None" if none applies. 
      You should select exclusively from these relation options and not introduce other relationships.
    }\\
    \texttt{Please structure your response as follows:\\
    Response::: \\
    Explanation: (your analysis of the semantic relation between two concepts). \\
    Answer: (state the relation explicitly, e.g. "ConceptA is [relation] ConceptB") \\}
    
    \texttt{
    \textbf{\%User Message:} \\
      Examine the relation between the following two concepts, each described below: \\
      $\ast$ ConceptA: labeled as "\{child\_label\}", described as "\{child\_description\}". \\
      $\ast$ ConceptB: labeled as "\{parent\_label\}", described as "\{parent\_description\}". \\
    }
    \textbf{\texttt{Response:::}}
  }
}}

\section{Evaluation \label{sec:evaluation}}

We assess the quality of WiKC from intrinsic and extrinsic perspectives. Data, code, prompts, and resources are all available online\footnote{\url{https://github.com/peng-yiwen/WiKC}}.

\textbf{Intrinsic evaluation.} We verified the inclusion of the 40 upper-level classes of YAGO 4.5\footnote{\url{https://yago-knowledge.org/data/yago4.5/design-document.pdf}} in WiKC. 
There are only two exceptions: {\it yago:Gender}, which can be represented by the property {\it sex or gender} (P21), and {\it schema:Taxon}, which is a specific class of the biological domain. 
Following YAGO 4.5 \cite{suchanek2024yago}, we evaluate WiKC in terms of three key criteria: complexity, conciseness, and understandability (fraction of classes having labels and descriptions).
The statistics are given in Table \ref{tab:statistics}. As expected, WiKC is much simpler and much more concise than Wikidata taxonomy. Compared to WiKC, Wikidata taxonomy has a factor higher than 200 in the number of classes, and a factor higher than 10 in the average number of paths from an instance to the root class {\it entity} (Q35120). 

\begin{table}[h]
\centering
\small 
\caption{Quality Measures}
\label{tab:statistics}
\begin{tabular}{llcc}
\hline
\textbf{Criterion} & \textbf{Metric} & \textbf{Wikidata}  & \textbf{WiKC} \\
\hline
\multirow{5}{*}{Complexity} 
 & Classes & 4.1M & \textbf{17k} \\
 & Top-level classes & 38 & \textbf{16} \\
 & Links & 4.8M & \textbf{20k} \\
 & Depth & 20 & \textbf{13} \\
 & Average paths to root & 37 & \textbf{2.9} \\ \hline
 
\multirow{3}{*}{Conciseness} &  Cycles & 35 & \textbf{0} \\
 & Redundant links & 500k & \textbf{0} \\
 & Classes without instances & 3.9M & \textbf{0} \\ \hline
 
Understandability & Labels and descriptions & 78\% & \textbf{100\%} \\ \hline

\end{tabular}
\end{table}

\textbf{Extrinsic evaluation.} We further evaluate WiKC on a task of entity typing, i.e., the prediction of the classes of an entity. This is a crucial task for various downstream tasks, like entity alignment \cite{Huang2021CrossknowledgegraphEA} or entity linking \cite{Gupta2017EntityLV}. Our evaluation includes the direct classes of each instance as well as their ancestors in the taxonomy, in order to assess the inconsistency of some taxonomic paths.

We collect instances from the Wikidata dump based on the {\it instanceOf} (P31) or {\it occupation} (P106) relations, excluding scholarly articles (described in Section \S\ref{sec:taxon_extract}), ensuring each instance has a label, a description, and an English Wikipedia page (resulting in 7M instances). We retype these instances using WiKC by assigning instances to their nearest classes in the taxonomy. To avoid class distribution imbalance (e.g., the class {\it person} can have 2.6M cumulative instances), we limit each class to 1000 instances and randomly sample 100k samples overall, resulting in nearly 1M type statements per taxonomy.
We design a judge LLM\footnote{We use the same model \textit{Mixtral-8x7B-Instruct-v0.1}.} to verify the accuracy of type statements based on the context provided by an instance. For example, given the context: {\it $\ast$Paris$\ast$ is described as the capital of France}, the LLM judges if the statement {\it $\ast$Paris$\ast$ is a [city or town], which means 'large human settlement'} is True or False. In this case, the class within brackets can be any ancestor of  {\it city} (Q515), the direct class of  {\it Paris} (Q90).

Table \ref{tab:results} demonstrates the accuracy of entity typing across different depths of the taxonomy on Wikidata and WiKC, where depth refers to the shortest distance from the root class {\it entity}. The results show that WiKC consistently outperforms Wikidata across all depth ranges. WiKC shows significant accuracy gains at deeper levels (depth 10 or more), suggesting that WiKC has resolved many inconsistency issues in the lower levels of the Wikidata taxonomy. The fact that accuracy is higher at a deeper level (depth 5 or more) compared to a shallow level on WiKC can be explained by the fact that more specific types are easier for LLMs to judge. For example, it is easier to classify {\it Motokazu Mori} (Q75688679) as a {\it poet} (Q49757) (depth 9) than as a {\it corporate body} (Q106668099) (depth 2).

\begin{table}[h]
\centering
    \caption{Accuracy of entity typing on Wikidata and WiKC.}
    \label{tab:results}
    \begin{tabular}{ccccc}
        \toprule
        \textbf{Depth} & \textbf{[0, 5)} & \textbf{[5, 10)} & \textbf{[10, $\infty$)} & \textbf{Macro} \\
        \midrule
        Wikidata & 41\% & 47\% & 37\% & 43\% \\ 
        WiKC & \textbf{67\%} & {\bf 76\%} & \textbf{75\%} & {\bf 70\%} \\ 
        \bottomrule
    \end{tabular}
\end{table}

\textbf{Discussion.} We here discuss some limits of our work.

(1) Problems with the LLM. The LLM might hallucinate by producing responses in conflict with the input prompt. For example, when checking the link {\it coke plant} (Q905318) → {\it coke} (Q192795), the LLM generates a response where {\it coke} (Q192795) is a subclass of {\it coke plant's product}, creating a new class instead of respecting the input. It should generate {\it none} if no appropriate relation is found between the two classes, rather than hallucinating a non-existent class in Wikidata. Additionally, the LLM can exhibit inconsistency between the explanation part and the answer part, although this happens rarely. For instance, in the explanation part, {\it input device} (Q864114) is analyzed as one type of {\it physical interface} (Q64830866) that {\it specifically provides data and signals to an information processing system}. But the answer incorrectly orders the classes, stating that {\it physical interface} is a subclass of {\it input device}. 

(2) Applications to downstream tasks. Even though we perform both intrinsic and extrinsic evaluations of WiKC, we cut away large parts of the Wikidata taxonomy (from about 4M to 17k classes). Therefore, it is also important to assess the knowledge coverage of WiKC and its usefulness for downstream tasks. 

\section{Conclusion\label{sec:conclude}}
In this paper, we propose WiKC, a cleaned version of Wikidata taxonomy, generated by an automated process combining zero-shot prompting on an open-source LLM and graph mining approaches. The objective is to address several known limits of Wikidata taxonomy, such as inaccurate taxonomic paths, redundancy across classes,  complexity, and ambiguity between instances and classes. Our approach consists of cutting irrelevant links, resolving reversed links, reducing transitive links, merging equivalent classes, rewiring links upon reconfirmation of LLM, and filtering out non-informative or rare classes. The experimental results show the improved accuracy and conciseness of WiKC compared to the original taxonomy of Wikidata.
In addition, we provide a mapping file from WiKC to Wikidata, encouraging the reuse of WiKC in various downstream tasks, such as entity recognition \cite{fetahu2023multiconer}, entity linking \cite{barba2022extend} and entity summarization \cite{chisholm2017learning}, to further validate its reliability and its coverage of general knowledge. 

For future work, we consider directions for exploring other open-source LLMs to clean and evaluate taxonomies based on our proposed pipeline, and investigating the trustworthiness of these LLMs in the taxonomy refinement task. It is also valuable to share this approach with the Wikidata community to further check its feasibility and help alleviate the burden of manual taxonomy cleaning \footnote{\url{https://www.wikidata.org/wiki/Wikidata:WikiProject\_Ontology/Cleaning\_Task\_Force}}.

%
%
%
\newpage
\bibliographystyle{splncs04}
\bibliography{biblio}

\begin{thebibliography}{10}
\providecommand{\url}[1]{\texttt{#1}}
\providecommand{\urlprefix}{URL }
\providecommand{\doi}[1]{https://doi.org/#1}

\bibitem{abian2018inconsist}
Abi{\'a}n, D., Guerra, F., Mart{\'\i}nez-Romanos, J., Trillo-Lado, R.: Wikidata
  and {DBpedia}: a comparative study. In: Semantic Keyword-Based Search on
  Structured Data Sources: Third International KEYSTONE Conference, IKC 2017,
  Gda{\'n}sk, Poland, September 11-12, 2017, Revised Selected Papers and COST
  Action IC1302 Reports 3. pp. 142--154. Springer (2018)

\bibitem{aly2019every}
Aly, R., Acharya, S., Ossa, A., K{\"o}hn, A., Biemann, C., Panchenko, A.: Every
  child should have parents: a taxonomy refinement algorithm based on
  hyperbolic term embeddings. arXiv preprint arXiv:1906.02002  (2019)

\bibitem{auer2007dbpedia}
Auer, S., Bizer, C., Kobilarov, G., Lehmann, J., Cyganiak, R., Ives, Z.:
  {DBpedia}: A nucleus for a web of open data. In: international semantic web
  conference. pp. 722--735. Springer (2007)

\bibitem{babaei2023llms4ol}
Babaei~Giglou, H., D’Souza, J., Auer, S.: Llms4ol: Large language models for
  ontology learning. In: International Semantic Web Conference. pp. 408--427.
  Springer (2023)

\bibitem{barba2022extend}
Barba, E., Procopio, L., Navigli, R.: Extend: Extractive entity disambiguation.
  In: Proceedings of the 60th Annual Meeting of the Association for
  Computational Linguistics (Volume 1: Long Papers). pp. 2478--2488 (2022)

\bibitem{brasileiro2016applying}
Brasileiro, F., Almeida, J.P.A., Carvalho, V.A., Guizzardi, G.: Applying a
  multi-level modeling theory to assess taxonomic hierarchies in {Wikidata}.
  In: Proceedings of the 25th international conference companion on World Wide
  Web. pp. 975--980 (2016)

\bibitem{carta2023iterative}
Carta, S., Giuliani, A., Piano, L., Podda, A.S., Pompianu, L., Tiddia, S.G.:
  Iterative zero-shot llm prompting for knowledge graph construction. arXiv
  preprint arXiv:2307.01128  (2023)

\bibitem{chen2023prompting}
Chen, B., Yi, F., Varr{\'o}, D.: Prompting or fine-tuning? a comparative study
  of large language models for taxonomy construction. In: 2023 ACM/IEEE
  International Conference on Model Driven Engineering Languages and Systems
  Companion (MODELS-C). pp. 588--596. IEEE (2023)

\bibitem{chisholm2017learning}
Chisholm, A., Radford, W., Hachey, B.: Learning to generate one-sentence
  biographies from wikidata. In: Proceedings of the 15th Conference of the
  European Chapter of the Association for Computational Linguistics: Volume 1,
  Long Papers. pp. 633--642 (2017)

\bibitem{fetahu2023multiconer}
Fetahu, B., Chen, Z., Kar, S., Rokhlenko, O., Malmasi, S.: Multiconer v2: a
  large multilingual dataset for fine-grained and noisy named entity
  recognition. In: Findings of the Association for Computational Linguistics:
  EMNLP 2023. pp. 2027--2051 (2023)

\bibitem{funk2023towards}
Funk, M., Hosemann, S., Jung, J.C., Lutz, C.: Towards ontology construction
  with language models. arXiv preprint arXiv:2309.09898  (2023)

\bibitem{Gupta2017EntityLV}
Gupta, N., Singh, S., Roth, D.: Entity linking via joint encoding of types,
  descriptions, and context. In: Conference on Empirical Methods in Natural
  Language Processing (2017),
  \url{https://api.semanticscholar.org/CorpusID:28784495}

\bibitem{Huang2021CrossknowledgegraphEA}
Huang, H., Li, C., Peng, X., He, L., Guo, S., Peng, H., Wang, L., Li, J.:
  Cross-knowledge-graph entity alignment via relation prediction. Knowl. Based
  Syst.  \textbf{240},  107813 (2021),
  \url{https://api.semanticscholar.org/CorpusID:245304392}

\bibitem{malandri2021taxoref}
Malandri, L., Mercorio, F., Mezzanzanica, M., Nobani, N.: Taxoref: Embeddings
  evaluation for ai-driven taxonomy refinement. In: Joint European Conference
  on Machine Learning and Knowledge Discovery in Databases. pp. 612--627.
  Springer (2021)

\bibitem{o2017restructuring}
O'Hara, T.D., Hugall, A.F., Thuy, B., St{\"o}hr, S., Martynov, A.V.:
  Restructuring higher taxonomy using broad-scale phylogenomics: the living
  ophiuroidea. Molecular phylogenetics and evolution  \textbf{107},  415--430
  (2017)

\bibitem{ponzetto2009large}
Ponzetto, S.P., Navigli, R., et~al.: Large-scale taxonomy mapping for
  restructuring and integrating wikipedia. In: IJCAI. vol.~9, pp. 2083--2088
  (2009)

\bibitem{dbpediaInconsist2012}
Sheng, Z., Wang, X., Shi, H., Feng, Z.: Checking and handling inconsistency of
  {DBpedia}. In: Web Information Systems and Mining: International Conference,
  WISM 2012, Chengdu, China, October 26-28, 2012. Proceedings. pp. 480--488.
  Springer (2012)

\bibitem{shenoy2022study}
Shenoy, K., Ilievski, F., Garijo, D., Schwabe, D., Szekely, P.: A study of the
  quality of {Wikidata}. Journal of Web Semantics  \textbf{72},  100679 (2022)

\bibitem{suchanek2024yago}
Suchanek, F.M., Alam, M., Bonald, T., Chen, L., Paris, P.H., Soria, J.: Yago
  4.5: A large and clean knowledge base with a rich taxonomy. In: Proceedings
  of the 47th International ACM SIGIR Conference on Research and Development in
  Information Retrieval. pp. 131--140 (2024)

\bibitem{sun2024large}
Sun, Y., Xin, H., Sun, K., Xu, Y.E., Yang, X., Dong, X.L., Tang, N., Chen, L.:
  Are large language models a good replacement of taxonomies? arXiv preprint
  arXiv:2406.11131  (2024)

\bibitem{vrandevcic2014wikidata}
Vrande{\v{c}}i{\'c}, D., Kr{\"o}tzsch, M.: Wikidata: a free collaborative
  knowledge base. Communications of the ACM  \textbf{57}(10),  78--85 (2014)

\bibitem{wei2022chain}
Wei, J., Wang, X., Schuurmans, D., Bosma, M., Xia, F., Chi, E., Le, Q.V., Zhou,
  D., et~al.: Chain-of-thought prompting elicits reasoning in large language
  models. Advances in neural information processing systems  \textbf{35},
  24824--24837 (2022)

\bibitem{zeng2024chain}
Zeng, Q., Bai, Y., Tan, Z., Feng, S., Liang, Z., Zhang, Z., Jiang, M.:
  Chain-of-layer: Iteratively prompting large language models for taxonomy
  induction from limited examples. arXiv preprint arXiv:2402.07386  (2024)

\end{thebibliography}
%




\end{document}